\documentclass[journal]{IEEEtran}
\usepackage{caption}
\usepackage{subcaption}
\usepackage{amsmath,amsfonts}
\usepackage{rotating}
\usepackage{algorithmic}
\usepackage{algorithm}
\usepackage{array}
\usepackage{textcomp}
\usepackage{stfloats}
\usepackage{url}
\usepackage{verbatim}
\usepackage{graphicx}
\usepackage{bm}
\usepackage{amssymb}
\usepackage{graphics}
\usepackage{graphicx}
\usepackage{color}
\usepackage{xcolor}
\usepackage{colortbl}
\usepackage{amsmath}
\usepackage{mathtools}
\usepackage{amsthm}
\usepackage{blindtext}
\usepackage{textcomp}
\usepackage{physics}
\usepackage{lscape}
\AfterEndEnvironment{strip}{\leavevmode}
\usepackage[numbers]{natbib}
\begin{document}

\title{Graph Embedded Intuitionistic Fuzzy Random Vector Functional Link Neural Network for \\Class Imbalance Learning\\}

\author{M.A. Ganaie{$^\dagger$}, M. Sajid{$^\dagger$}, A.K. Malik, M. Tanveer{$^*$}, ~\IEEEmembership{Senior Member,~IEEE,} for the Alzheimer’s Disease Neuroimaging Initiative 
\thanks{ \noindent $^*$ Corresponding Author. $^\dagger$ denotes the authors have equal contributions.\\
   M.A. Ganaie is with the Department of Computer Science and Engineering, Indian Institute of Technology Ropar, Punjab, 140001, India (e-mail: mudasir@iitrpr.ac.in). M. Sajid, A.K. Malik, and M. Tanveer are with the Department of Mathematics, Indian Institute of Technology Indore, Simrol, Indore, 453552, India (e-mail: phd2101241003@iiti.ac.in, phd1801241003@iiti.ac.in, mtanveer@iiti.ac.in). This study used data from the Alzheimer's Disease Neuroimaging Initiative (ADNI) (adni.loni.usc.edu).\\
   Note: This article is accepted in IEEE Transactions on Neural Networks and Learning Systems and has supplementary downloadable material available at https://doi.org/10.1109/TNNLS.2024.3353531 provided by the authors. Digital Object Identifier 10.1109/TNNLS.2024.3353531}}

\markboth{  }%
{Shell \MakeLowercase{\textit{et al.}}: A Sample Article Using IEEEtran.cls for IEEE Journals}

\maketitle
\begin{abstract}
The domain of machine learning is confronted with a crucial research area known as class imbalance learning, which presents considerable hurdles in precise classification of minority classes. This issue can result in biased models where the majority class takes precedence in the training process, leading to the underrepresentation of the minority class. The random vector functional link (RVFL) network is a widely used and effective learning model for classification due to its good generalization performance and efficiency. However, it suffers when dealing with imbalanced datasets. To overcome this limitation, we propose a novel graph embedded intuitionistic fuzzy RVFL for class imbalance learning (GE-IFRVFL-CIL) model incorporating a weighting mechanism to handle imbalanced datasets. The proposed GE-IFRVFL-CIL model offers plethora of benefits: $(i)$ leveraging graph embedding to preserve the inherent topological structure of the datasets, $(ii)$ employing intuitionistic fuzzy theory to handle uncertainty and imprecision in the data, $(iii)$ and the most important, it tackles class imbalance learning. The amalgamation of a weighting scheme, graph embedding, and intuitionistic fuzzy sets leads to the superior performance of the proposed models on KEEL benchmark imbalanced datasets with and without Gaussian noise. Furthermore, we implemented the proposed GE-IFRVFL-CIL on the ADNI dataset and achieved promising results, demonstrating the model's effectiveness in real-world applications. The proposed GE-IFRVFL-CIL model offers a promising solution to address the class imbalance issue, mitigates the detrimental effect of noise and outliers, and preserves the inherent geometrical structures of the dataset.
\end{abstract} 


\section{Introduction}
\IEEEPARstart{T}{he} ability of artificial neural networks (ANNs) to approximate nonlinear mappings is the primary reason for their huge success in many disciplines among numerous machine learning approaches \cite{abiodun2018state}. 
ANNs have demonstrated success in various fields such as rainfall forecasting \cite{luk2001application}, clinical medicine \cite{baxt1995application}, stock market predictions \cite{dase2010application}, solving differential equations \cite{lagaris1998artificial}, brain age prediction \cite{tanveer2023deep} and so on.  

The gradient descent (GD) method, an iterative process, is one of the most often used techniques for optimizing the cost function to train ANNs. In the GD-based technique, the difference between the real output and the anticipated output of the model backpropagates in an effort to optimize the weights and biases of the model. This iterative strategy has a number of underlying issues, such as being time-consuming having a tendency to converge to local rather than global optima \cite{gori1992problem}, being extremely sensitive to the choice of the learning rate and the point of initialization of the iteration.    
 
Randomized neural networks (RNNs) \cite{schmidt1992feed} were proposed to avoid the pitfalls of GD-based neural networks. In RNN, some network parameters are fixed during the training period, and only the parameters of the output layer are calculated via the closed-form solution \cite{suganthan2021origins}. RVFL network \cite{pao1994learning}, and extreme learning machine (ELM) \cite{huang2006extreme} are among the prominent RNNs. The distinctive feature setting RVFL apart from other RNNs lies in its establishment of direct linkages between the input and output layers.
The weights and biases within the hidden layer of the RVFL are randomly generated and kept fixed throughout the training phase. The output parameters, namely direct link weights and the weights connecting the hidden layer to the output layer, are analytically calculated using the Pseudo-inverse or the least-square method.
The incorporation of direct links in the RVFL has been observed to significantly improve the learning performance by functioning as a regularization for the randomization \cite{zhang2016comprehensive}. 
Furthermore, the thinner topology of the RVFL, compared to the ELM, aids in reducing its complexity and enhancing attractiveness in alignment with Occam's principle and probably approximately correct (PAC) learning theory \cite{kearns1994introduction, shi2021random}. The RVFL offers fast training speed as well as universal approximation ability \cite{igelnik1995stochastic}. 

Numerous variants of the original RVFL model have been developed to
improve its generalization performance \cite{malik2023random, Ganaie2022}.
The RVFL model converts the original features to randomized features, which makes RVFL unstable. A sparse autoencoder with $l1$-norm regularization was employed in RVFL (SP-RVFL) \cite{zhang2019unsupervised}. The SP-RVFL deals with the issue of instability caused by randomization and learns the network parameters more appropriately than the traditional RVFL.
In \cite{zhang2020new}, the authors presented two models, namely RVFL+ (incorporating RVFL with learning using privileged information (LUPI)) and  KRVFL+ (Kernel-based RVFL+). During the training phase, RVFL+ benefits from privileged data in addition to the training data. KRVFL+ manages nonlinear interactions between higher dimensional input and output vectors along with the privileged information. RVFL+ still has the same challenge as RVFL networks: it is challenging to construct the ideal number of hidden nodes. The number of hidden nodes has a major impact on the efficacy of the network's learning process \cite{li2017bayesian}. \citet{dai2022incremental} developed an algorithm called Incremental RVFL+ (IRVFL+) that is constructive in nature. The IRVFL+ network continually enlarges the hidden nodes of the network to resemble the output. In \cite{chakravorti2020non}, kernel-based exponentially expanded RVFL (KERVFL) was put forth to avoid the search for an optimal number of hidden nodes by adding the kernel function to the RVFL model.

Certain data points in a dataset that possess qualities and attributes typically associated with one class or category but instead fall into a distinct class or category are known as outliers. The prediction accuracy of machine learning models is severely impacted by noisy data, outliers, and imbalanced datasets. The class imbalance (CI) problem refers to a scenario when there is a substantial disparity in the number of samples pertaining to a particular class in a given dataset compared to the remaining classes. Since the standard RVFL assigns a uniform weighting scheme to each sample while generating the optimal classifier, it is vulnerable to noise, outliers, and imbalanced issues despite its excellent computational efficiency and strong generalization capacity in balanced datasets.
The fuzzy theory has been effectively used to mitigate the detrimental effects of noise or outliers on machine learning models' performance \cite{rezvani2019intuitionistic}. 
The fuzzy approach defines a degree membership function by assisting the notion of the distance of data samples to the corresponding class center.
In \cite{ha2013support}, intuitionistic fuzzy (IF) membership, an extended version of fuzzy membership, was proposed.
The IF membership function assigns an IF score to each sample with the help of membership and nonmembership functions. Intuitionistic fuzzy RVFL (IFRVFL) \cite{malik2022alzheimer}
for datasets with noise and outliers, outperformed standard RVFL in terms of generalization performance; however, IFRVFL has not yet addressed the issue of imbalanced datasets.

The original RVFL disregards the geometrical relationship of the data while computing the final output parameters \cite{malik2023random}. Many improved versions of the RVFL have been proposed as remedies for the aforementioned problem. \citet{li2021discriminative} proposed a manifold learning-based variant of RVFL, namely discriminative manifold RVFL (DMRVFL). In order to effectively utilize the intraclass discriminative information and concurrently increase the distances between interclass samples, DMRVFL replaced the inflexible one-hot label matrix with a more flexible soft label matrix. On the one hand, DMRVFL enlarges the distance of the interclass samples. On the other hand, it makes the intraclass samples more compact.
 \citet{ganaie2020minimum} presented two multiclass classifiers, namely Class-Var-RVFL and Total-Var-RVFL, which utilize both the original and the projected randomized space's dispersion of training data to optimize output layer weights. 
 The former relies on intraclass variance minimization, while the latter employs total variance minimization.
 Recently, graph embedded intuitionistic fuzzy weighted RVFL (GE-IFWRVFL) \cite{MalikGraph2022} was proposed by incorporating subspace learning (SL) criteria within the graph embedding (GE) framework with RVFL. 
Although GE-IFWRVFL brings forth many benefits, such as it preserves the geometrical property of the dataset and 
 IF assists the model in handling outliers and noisy data samples; however, GE-IFWRVFL fails to deal with the imbalanced datasets.
%

In reality, patterns from the minority class can be more significant than patterns from other classes.
Conventional RVFL treats each sample uniformly irrespective of its belongingness to minority or majority classes, and as a result, RVFL is biased in favor of the majority class samples.
Thus, CI issues hinder the performance of the RVFL model. The improved fuzziness-based RVFL (IF-RVFL) \cite{cao2020improved} demonstrated its generalization performance in real-life liver disease imbalanced datasets. Class-specific weighted RVFL (CSWRVFL) \cite{sahani2019fpgaa} handles the dilemma of imbalanced datasets and carried out tests to identify the power quality disturbance. Although RVFL has made significant progress wherein IFRVFL \cite{malik2022alzheimer} deals with noise and outliers and CSWRVFL \cite{sahani2019fpgaa} and IF-RVFL \cite{cao2020improved} retains the topological relationship of the data, there has been relatively little improvement in dealing with imbalance datasets. To the best of our knowledge, there is currently no existing model within the RVFL family that effectively tackles the challenges associated with noise and outliers while also maintaining the inherent geometric structure of the data and simultaneously addressing the challenge of imbalanced datasets in a holistic manner.
To bridge this gap, we propose a novel graph embedded intuitionistic fuzzy random vector functional link network for class imbalance learning (GE-IFRVFL-CIL). The GE framework, IF membership functions and CI weighting enable the proposed GE-IFRVFL-CIL model to address the imbalance issue along with noise, outliers, and geometrical structure preservation of the dataset.
This paper's primary highlights are as follows:
\begin{enumerate}
    \item We propose a generic framework that can adapt to different criteria used to handle imbalance scenarios. To demonstrate the efficacy, the proposed GE-IFRVFL-CIL model incorporates a weighting scheme \cite{rezvani2022intuitionistic} to address the CI issue. The weighting scheme takes care of the minority class by assigning unit weightage to its samples, and the weights of the majority class are lowered by the ratio of the number of positive class samples to the number of negative class samples.
    \item To compute the output weights, the proposed GE-IFRVFL-CIL model integrates SL criteria, utilizing both intrinsic and penalty SL within the GE framework. Under the GE framework, linear discriminant analysis (LDA) \cite{martinez2001pca} and local Fisher discriminant analysis (LFDA) \cite{sugiyama2007dimensionality} are employed, yielding two alternative combinations of models. The GE, along with the utilization of a regularization term, seeks to maintain graph structural information in the projected space.   
    \item The IF membership technique is employed in the proposed GE-IFRVFL-CIL model to deal with noisy and outlier samples in the datasets. The degree of membership and non-membership are assigned to each sample in the IF concept to determine if it is a pure sample, noise, or outlier, and data samples are weighted accordingly.
    \item We demonstrate the efficacy of the proposed GE-IFRVFL-CIL models over the KEEL imbalanced datasets (with and without Gaussian noise) from various domains, and diagnosis the Alzheimer's disease using the ADNI dataset. Experiments confirmed that the proposed GE-IFRVFL-CIL-1 and GE-IFRVFL-CIL-2 models outperform numerous state-of-the-art models in the context of class imbalance learning.
\end{enumerate}
The subsequent sections of this paper are arranged as follows. Section \ref{Related_works} presents a concise introduction to the GE framework. Section \ref{proposed_work} provides the mathematical formulation of the proposed models, followed by a weighting mechanism for class imbalanced learning, and discusses LDA and LFDA under the GE framework. The detailed highlights and comparison of the proposed models with the baseline models are discussed in Section II of the Supplementary material. Experimental results are demonstrated in Section \ref{experiments} on KEEL datasets with and without Gaussian noise and on ADNI datasets to detect Alzheimer's disease. Section \ref{conclusion} includes a conclusion and some recommendations for future work. 
 \section{Preliminary Works}
\label{Related_works}
This section provides a brief overview of the GE framework \cite{4016549}. The RVFL structure, IFRVFL, and ELM networks are discussed in Section I of the Supplementary materials.
\subsection{Notations}
Let $\bigl\{(x_k,y_k)\vert\, x_k \in \mathbb{R}^p,\, y_k \in \mathbb{R}^q; k = 1, \hdots, N\bigl\}$ be the training set, where $p$ denotes the number of features in each input sample, and $q$ denotes the number of classes with $N$ training data samples. Let $X \in \mathbb{R}^{N \times p}$ be the input matrix and $Y \in \mathbb{R}^{N \times q}$ be the target matrix. $H \in \mathbb{R}^{N \times h_l}$ is the hidden layer matrix of RVFL and ELM obtained by transforming the input matrix with the help of randomly initialized weights and biases followed by the non-linear activation function $\Phi$, where $h_l$ is the number of hidden layer nodes.
\subsection{Graph Embedding (GE) \cite{4016549}}
\label{subsection:G_matrix_Graph}
\citet{4016549} proposed a GE algorithm that reformulates several dimensionality reduction methods within a unified framework and also aids in the design of new algorithms. The embedding process aims to preserve important structural information of the graph in the resulting vector space. In GE, for input dataset $X=\bigl\{x_k~\vert\ x_k \in \mathbb{R}^p,\; k = 1, \hdots,N\bigl\}$, the intrinsic graph, $\mathcal{G}^{int}=\{X,\Delta^{int}\}$, and the penalty graph $\mathcal{G}^{pen}=\{X,\Delta^{pen}\}$ are defined. The weights relating to the unique association between two vertices in $X$ are included in the similarity weight matrix $\Delta^{int} \in \mathbb{R}^{N \times N}$. Each component of the penalty weight matrix, $\Delta^{pen} \in \mathbb{R}^{N\times N}$, is the penalty matrix of $X$ that takes into account a particular relationship between the vertices of the graph. The graph embedding optimization problem is defined as follows:
\begin{align} 
\label{eqn:GE1}
\hat{v}=&\underset {Tr({v_{0}^{T}X^{T}\mathcal{U}X{v_{0}})} = c}{\rm argmin} \sum _{k\neq l} \left \|{  {v_{0}}^{T}  {x}_{k} -  {v_{0}}^{T}  {x}_{l} }\right \|_2^{2} \Delta_{kl}^{int} \notag \\=&\underset {Tr({ {v_{0}}^{T}  {X}^{T}  \mathcal{U}  {X}  {v_{0}})}= c}{\rm argmin} {  Tr({v_{0}}^{T}  {X}^{T}  {\mathcal{L}} {X} {v_{0}}) }. \end{align}
Here, the operator $Tr(\cdot)$ denotes the trace of a matrix, $v_{0}$ is the projection matrix, $\mathcal{L}=\mathcal{D}-\Delta^{int}\in \mathbb{R}^{N \times N}$ is the graph Laplacian matrix of the intrinsic graph $\mathcal{G}^{int}$ and the elements of the diagonal matrix $\mathcal{D}$ is defined as $\mathcal{D}_{kk}=\sum_{l}\Delta_{kl}^{int}$.  $\mathcal{U}=\mathcal{L}^{p}=\mathcal{D}^{p}-\Delta^{pen}$ is the Laplacian matrix of penalty graph $\mathcal{G}^{pen}$ or a diagonal matrix for normalizing scale, and $c$ is a constant value. The optimization problem (\ref{eqn:GE1}) boils down to a generalized eigenvalue problem \cite{chung1997spectral},
\begin{align}
  {G}_{i} {s} = \lambda  {G}_{p} {s},\end{align}
here, $G_{i}=X^{T}\mathcal{L}X$ and ${G}_{p}=X^{T}\mathcal{U}X$. It implies that the transformation matrix will be formed by the eigenvectors of matrix $G={G}_{p}^{-1}{G}_{i}$. The matrix $G$ takes into account data samples' intrinsic and penalty graph connections.
\section{The Proposed Graph Embedded Intuitionistic Fuzzy RVFL for Class Imbalance Learning (GE-IFRVFL-CIL)}
\label{proposed_work}
This section provides a detailed description of the proposed GE-IFRVFL-CIL model. We first define the generic mathematical framework of the
proposed model to handle the class imbalance issue. The proposed GE-IFRVFL-CIL model addresses the class imbalance issue by incorporating a weighting scheme based on the class imbalance ratio. The optimization process for calculating network output weights integrates subspace learning (SL) criteria, leveraging both intrinsic and penalty SL within the GE framework. To preserve the geometrical structure,  LDA \cite{martinez2001pca} and LFDA \cite{sugiyama2007dimensionality} are incorporated under the GE framework, along with a GE regularization parameter. Additionally, IF numbers are assigned to each training sample for handling noise and outliers. The optimization problem of the proposed GE-IFRVFL-CIL model is defined as follows:
\begin{align} 
\label{peq:1}
    \underset{\beta}{\text{minimize}}~~ &\frac{1}{2}\norm{\beta}_2^2+\frac{\theta}{2}\norm{G^{\frac{1}{2}}\beta}_2^2+\frac{C}{2}(d_+)\norm{S_+^{\frac{1}{2}}\xi_+}_2^2+ \nonumber\\
 &\frac{C}{2}(d_-)\norm{S_-^{\frac{1}{2}}\xi_-}_2^2,  \nonumber\\
    s.t.~~& Z_{+}\beta=Y_+-\xi_+ ~\text{and}~ Z_{-}\beta=Y_--\xi_-. 
\end{align}
Here, $\beta$ is the weight matrix connecting the hidden layer and the input layer to the output layer, $\theta$ is a graph regularization parameter, $G$ is the graph embedding matrix defined in Subsection \ref{subsection:G_matrix_Graph}, $S_-$ ($S_+$) is the diagonal matrix with IF scores along its diagonal for the negative (positive) class samples. $\xi_-$ ($\xi_+$) is the error of the negative (positive) class samples. Here, $d_-$ ($d_+$) is the weighting scheme of the negative (positive) class defined in Subsection \ref{subsection:weight}. $C \in \mathbb{R^+}$ is the regularization parameter to penalize the error variables.  $Z_{-}$ ($Z_{+}$) denotes the non-linear as well as linear projection of the negative (positive) class samples, defined as:
\begin{equation}
    Z_{-}=\left[\begin{array}{ll} X_{-} & H_{neg}\end{array}\right] ~\text{and}~ Z_{+}=\left[\begin{array}{ll} X_{+} & H_{pos}\end{array}\right],
\end{equation}
where $X_{-} ~(X_{+})$ is the negative (positive) class' input matrix and $H_{neg} ~(H_{pos})$ is the negative (positive) class' hidden layer matrix obtained by transforming $X_{-} ~(X_{+})$ with the help of randomly initialized weights
and biases followed by the non-linear activation function $\Phi$. $Y_-$ ($Y_+$) is the target matrix for the negative (positive) class samples. 

The objective function in \eqref{peq:1} comprises four key components: \\
1. Structural Risk Minimization (SRM): The first term aims to minimize the norm of the parameter $\beta$, adhering to the principle of SRM. \\
2. Graph Embedding Regularization: The second term allocates GE weights to $\beta$ and minimizes the resultant norm, promoting a structured relationship between samples.\\
3. Positive Class Enhancement: The third term is designed for the positive class and assigns IF weights to the positive class error. This component considers the CI by incorporating the $d_+$ (positive CI weights) to each sample, with the weights being higher for minority class samples.\\
4. Negative Class Enhancement: The fourth term is analogous to the third but pertains to the negative class. It assigns IF weights to the negative class error, incorporating the $d_-$ (negative CI weights), which favor minority class samples.
Hence, the introduction of CI weights ($d_+$ and $d_-$) effectively balances the influence of both majority and minority class samples during training, leading to improved learning compared to existing baseline models.

The Lagrangian of \eqref{peq:1} is given as follows:
\begin{align}
\label{peq:e1}
    L= & \frac{1}{2}\norm{\beta}_2^2+\frac{\theta}{2}\norm{G^{\frac{1}{2}}\beta}_2^2+  \frac{C}{2}(d_+)\norm{S_+^{\frac{1}{2}}\xi_+}_2^2+ \nonumber\\
 & \frac{C}{2}(d_-)\norm{S_-^{\frac{1}{2}}\xi_-}_2^2  
    -\alpha_+^T(Z_{+}\beta-Y_++\xi_+) - \nonumber\\ & \alpha^T_-(Z_{-}\beta-Y_-+\xi_-), 
\end{align}
where $\alpha_+$ and $\alpha_-$ are the Lagrange multipliers. Let
\begin{align}
    l_{+}= C \times d_+ \text{  and   } l_{-}= C\times d_-.
\end{align}
Rewriting \eqref{peq:e1}, we have
\begin{align}
\label{peqn:30}
     L=&\frac{1}{2}\norm{\beta}^2_2+\frac{\theta}{2}\norm{G^{\frac{1}{2}}\beta}^2_2+\frac{l_{+}}{2}\norm{S_+^{\frac{1}{2}}\xi_+}^2_2+\frac{l_{-}}{2}\norm{S_-^{\frac{1}{2}}\xi_-}^2_2   \nonumber\\
 &
    -\begin{bmatrix}
    \alpha^T_+ & \alpha^T_- 
    \end{bmatrix}
    \Bigg( \begin{bmatrix}
           Z_{+}\\Z_{-}
           \end{bmatrix} \beta - \begin{bmatrix}
                                    Y_+\\Y_-
                                \end{bmatrix}+   \begin{bmatrix}
                                    \xi_+\\\xi_-
                                \end{bmatrix} \Bigg).
\end{align}
 By applying K.K.T. conditions to  \eqref{peqn:30}, we get
 \begin{align}
     &\beta+\theta G \beta -(Z^T_{+}\alpha_+ + Z^T_{-}\alpha_-)=0, \label{eqn:kkt1} \\ 
    &l_{+}S_+\xi_+-\alpha_+=0, \label{eqn:kkt2} \\
    &l_{-}S_-\xi_--\alpha_-=0, \label{eqn:kkt3}\\
    &Z_{+}\beta-Y_++\xi_+=0, \label{eqn:kkt4} \\
    &Z_{-}\beta-Y_-+\xi_-=0. \label{eqn:kkt5} 
 \end{align}
 Rewriting the equations \eqref{eqn:kkt1}, \eqref{eqn:kkt2} and \eqref{eqn:kkt3}, we have
 \begin{align}
 \label{peqn:37}
     &(I+\theta G)\beta=(Z^T_{+}\alpha_+ + Z^T_{-}\alpha_-),  \\
     \label{peqn:38}
    &l_{+}S_+\xi_+=\alpha_+, \\
    \label{peqn:39}
    &l_{-}S_-\xi_-=\alpha_-,
 \end{align}
where $I$ is an identity matrix of conformal dimensions. Using \eqref{peqn:38} and \eqref{peqn:39} in \eqref{peqn:37}, we get
 \begin{align}
   \label{peqn:40}
     (I+\theta G)\beta=&l_{+}Z^T_{+}S_+\xi_+ + l_{-}Z^T_{-}S_-\xi_- ,\\
       \label{peqn:41}
      (I+\theta G)\beta= &l_{+}Z^T_{+}S_+(Y_+-Z_{+}\beta) + \nonumber\\
 & ~l_{-}Z^T_{-}S_-(Y_--Z_{-}\beta),
       \end{align}
       \begin{align}
        \label{peqn:42}
     & (I+\theta G +l_{+}Z^T_{+}S_+Z_{+} + l_{-}Z^T_{-}S_-Z_{-})\beta\nonumber\\
 & = l_{+}Z^T_{+}S_+Y_+ + l_{-}Z^T_{-}S_-Y_-,
 \end{align}
 Multiply $(\frac{1}{l_{+}}+\frac{1}{l_{-}})$ to \eqref{peqn:42}, we have
 \begin{align}
 &\Bigg(\frac{1}{l_{+}}+\frac{1}{l_{-}}\Bigg) \Bigg(I+\theta G +l_{+}Z^T_{+}S_+Z_{+} + l_{-}Z^T_{-}S_-Z_{-}\Bigg)\beta  \nonumber\\
 & = \Bigg(\frac{1}{l_{+}}+\frac{1}{l_{-}}\Bigg) \Bigg(l_{+}Z^T_{+}S_+Y_+ + l_{-}Z^T_{-}S_-Y_- \Bigg),
 \end{align}
 \begin{align}
 &\Bigg( \Big(\frac{1}{l_{+}}+\frac{1}{l_{-}}\Big) \Big(I+\theta G\Big) + Z^T_{+}S_+Z_{+} + \frac{l_{-}}{l_{+}}Z^T_{-}S_-Z_{-}   \nonumber\\& +\frac{l_{+}}{l_{-}}Z^T_{+}S_+Z_{+} + Z^T_{-}S_-Z_{-} \Bigg) \beta = Z^T_{+}S_+Y_+ + \nonumber\\
 & \frac{l_{-}}{l_{+}}Z^T_{-}S_-Y_- + \frac{l_{+}}{l_{-}}Z^T_{+}S_+Y_+ + Z^T_{-}S_-Y_-,
 \end{align}
 \begin{align}
 &\beta=\Bigg( \Big(\frac{1}{l_{+}}+\frac{1}{l_{-}}\Big) \Big(I+\theta G\Big) + Z^T_{+}S_+Z_{+} + \frac{l_{-}}{l_{+}}Z^T_{-}S_-Z_{-}   \nonumber\\
 & + \frac{l_{+}}{l_{-}}Z^T_{+}S_+Z_{+} + Z^T_{-}S_-Z_{-} \Bigg)^{-1} \Bigg(Z^T_{+}S_+Y_+ + \frac{l_{-}}{l_{+}}Z^T_{-}S_-Y_- \nonumber\\
 & + \frac{l_{+}}{l_{-}}Z^T_{+}S_+Y_+ + Z^T_{-}S_-Y_-\Bigg),
 \end{align}
 \begin{align}
& \beta=\Bigg( \Big(\frac{1}{l_{+}}+\frac{1}{l_{-}}\Big) \Big(I+\theta G\Big) + (1+\frac{l_{+}}{l_{-}})Z^T_{+}S_+Z_{+}~ + \nonumber\\
 &(1+ \frac{l_{-}}{l_{+}})Z^T_{-}S_-Z_{-}   \Bigg)^{-1}\Bigg((1+\frac{l_{+}}{l_{-}})Z^T_{+}S_+Y_+ ~+\nonumber \\
 & ( 1+\frac{l_{-}}{l_{+}})Z^T_{-}S_-Y_- \Bigg),
\end{align}
 \begin{align}
 \label{Final_weight_beta}
&\beta=\Bigg( \big(\frac{1}{l_{+}}+\frac{1}{l_{-}}\big) \big(I+\theta G\big) +[Z_{+}^T,~ Z_{-}^T]\nonumber \\
 &
 \begin{bmatrix}
 (1+\frac{l_{+}}{l_{-}})S_+ &\bm{0}\\
 \bm{0}    & (1+\frac{l_{-}}{l_{+}})S_-
 \end{bmatrix}
 \begin{bmatrix}
 Z_{+}\\Z_{-}
 \end{bmatrix}\Bigg)^{-1}  \nonumber\\
 &
 \begin{bmatrix}
 Z_{+}^T &Z_{-}^T
 \end{bmatrix}
 \begin{bmatrix}
 (1+\frac{l_{+}}{l_{-}})S_+&\bm{0}\\
 \bm{0}&    ( 1+\frac{l_{-}}{l_{+}})S_- 
 \end{bmatrix}\begin{bmatrix}
 Y_+\\Y_-
 \end{bmatrix}.
 \end{align}
Using equation \eqref{Final_weight_beta}, we find the output layer's weight matrix.
 
Consequent Sections discuss LDA and LFDA under the GE framework and the class imbalance weighting scheme.
\subsection{Class Imbalance Weighting Scheme \cite{rezvani2022intuitionistic}}
\label{subsection:weight}
Let $l_{n}$ ($l_{p}$) denote the number of samples of the negative (positive) class.
    \begin{align}
d_{+}&= 1, ~\text{if}~x_i~\text{is in the positive class,}\\
 d_{-}&=\frac{l_p}{l_n},  ~\text{if}~x_i~\text{is in the negative class}.
    \end{align}
In this weighting method, the sample of the minority class is given unit weights, and the weight of the majority class is 
reduced by a factor equal to the proportion of positive class 
samples to negative class samples.

In this way, the proposed GE-IFRVFL-CIL model reduces the influence of majority class samples during training by correspondingly lowering their weights, effectively balancing sample weights between positive and negative classes. This CI weighting strategy empowers the minority class in training while maintaining the weights between both classes to some extent. This leads to improved parameter learning in the proposed GE-IFRVFL-CIL compared to baseline models. 
\subsection{LDA and LFDA under the GE Framework \cite{4016549}}
\label{subsection:graph}
Both the intrinsic as well as penalty graphs are based on the concatenated matrix $Z$, {\em{i.e.,}} $\mathcal{G}^{int}=\{Z,\Delta^{int}\}$ and $\mathcal{G}^{pen}=\{Z,\Delta^{pen}\}$, respectively. Therefore, ${G}_{i}=Z^{T}\mathcal{L}Z$ and ${G}_{p}=Z^{T}\mathcal{U}Z$. Weights for intrinsic and penalty graphs for the LDA and LFDA are given as follows:
\begin{itemize}
    \item \textbf{{Linear discriminant analysis (LDA)}:} The intrinsic and penalty graph weights for the LDA model are as follows:
    \begin{align}
        \label{LDAgraph1}
        \Delta^{int}_{ij} &= \left\{\begin{array}{ll} \frac{1}{l_{c_i}}, & \text{if}~~ c_i = c_j ,\vspace{3mm} \\ 
0, & \text{otherwise}. \end{array}\right.\\
        \label{LDAgraph11}
        \Delta^{pen}_{ij} &= \left\{\begin{array}{ll} \frac{1}{N}-\frac{1}{l_{c_i}}, & \text{if}~~ c_i = c_j ,\vspace{3mm} \\ 
\frac{1}{N}, & \text{otherwise}. \end{array}\right.
    \end{align}
    Here, $l_{c_i}$ denotes the number of samples in the class $c_i$.
     \item \textbf{{Local Fisher discriminant analysis (LFDA)}:} LFDA model assigns intrinsic and penalty graph weights as follows:
    \begin{align}
        \label{LFDAgraph1}
        \Delta^{int}_{ij} &= \left\{\begin{array}{ll} \frac{\eta_{ij}}{l_{c_i}}, & \text{if}~~ c_i = c_j ,\vspace{3mm} \\ 
0, & \text{otherwise}. \end{array}\right.\\
        \label{LFDAgraph11}
        \Delta^{pen}_{ij} &= \left\{\begin{array}{ll} \eta_{ij}\bigg(\frac{1}{N}-\frac{1}{l_{c_i}}\bigg), & \text{if}~~ c_i = c_j ,\vspace{3mm} \\ 
\frac{1}{N}, & \text{otherwise}. \end{array}\right.
    \end{align}
   Here, $\eta_{ij}=exp(-\frac{\norm{z_j-z_i}^2}{2\sigma^2})$, where $z_j, ~z_i\in Z$ and $\sigma$ is the scaling parameter.
   The similarity between $z_j~ \text{and} ~z_i$ in the matrix $Z$ is measured by $\eta_{ij}$.
\end{itemize}
To effectively contend with the challenge of imbalanced datasets, we used a weighting scheme in conjunction with two distinct graph embedding techniques, namely LDA and LFDA. From now onwards, GE-IFRVFL-CIL with LDA structure is referred to as GE-IFRVFL-CIL-1, and GE-IFRVFL-CIL with LFDA structure is referred to as GE-IFRVFL-CIL-2. 
 \section{Computational Complexity of the Proposed GE-IFRVFL-CIL Models}
In the solution provided for GE-IFRVFL-CIL in \eqref{Final_weight_beta}, and following the methodology in \cite{iosifidis2015graph}, the computation of the graph embedding matrix $G$ considers both intrinsic and penalty graph structures. The time complexity associated with this process is $\mathcal{O}((p+h_l)^3 + (p+h_l)^2N)$. As per \cite{rezvani2019intuitionistic}, the intuitionistic fuzzy score matrix for each sample has time complexity $\mathcal{O}(N)$ and the imbalance ratio assignment requires $\mathcal{O}(N)$ complexity. Following the standard procedure for determining the time complexity of matrix inversion and multiplication of matrix, the time complexity of a matrix inversion and the four matrix multiplication in \eqref{Final_weight_beta} requires a time complexity of $\mathcal{O}((p+h_l)^3 + (p+h_l)^2N)$. Thus,
the resultant time complexity of the proposed GE-IFRVFL-CIL model is equal
to $\mathcal{O}((p+h_l)^3 + (p+h_l)^2N)+\mathcal{O}(N)+\mathcal{O}(N)+\mathcal{O}((p+h_l)^3 + (p+h_l)^2N) \thickapprox \mathcal{O}((p+h_l)^3 + (p+h_l)^2N)$.
 \section{Experiments and Results}
\label{experiments}
The experiments are executed on a computing system possessing MATLAB R2017b software, an Intel(R) Xeon(R) CPU E5-2697 v4 processor operating at 2.30 GHz with 128-GB Random Access Memory (RAM), and a Windows-10 operating platform.  
 To generate IF weights in IFKRR \cite{hazarika2021intuitionistic}, IFRVFL \cite{malik2022alzheimer}, GE-IFWRVFL \cite{MalikGraph2022} and IFTWSVM \cite{rezvani2019intuitionistic} models, the Gaussian Kernel function is employed and is defined as: $K(x_1,x_2)=exp(-\frac{\norm{x_1-x_2}^2}{\mu^2})$, where $\mu$ is the kernel parameter. The experimental procedure and hyperparameter settings are discussed in Section III of the Supplementary material.
 \subsection{Evaluation on KEEL benchmark datasets} 
We demonstrate the efficacy of our proposed models in class imbalance learning by leveraging $28$ standard benchmark datasets sourced from the KEEL imbalanced dataset repository \cite{derrac2015keel}. These datasets encompass diverse domains and imbalance ratios (IR). Specifically, we strategically select $14$ small datasets with sample sizes below $500$ and $14$ medium-sized datasets with sample sizes exceeding $500$. For further details, including statistical insights and dataset-specific imbalance ratios, please refer to Table II of Supplementary material.
We compare the proposed GE-IFRVFL-CIL-1 and GE-IFRVFL-CIL-2 models with various machine learning models, namely intuitionistic fuzzy twin support vector machine (IFTWSVM) \cite{rezvani2019intuitionistic}, RVFL \cite{pao1994learning}, intuitionistic fuzzy kernel ridge regression (IFKRR) \cite{hazarika2021intuitionistic}, ELM \cite{huang2006extreme}, IFRVFL \cite{malik2022alzheimer} and GE-IFWRVFL \cite{MalikGraph2022}. The area under the curve (AUC) from Table \ref{tab:GE-IFRVFL-CIL} shows the performance of the models in classifying data. We follow five metrics, namely average AUC, ranking scheme, Friedman test, Nemenyi post hoc test, and win-tie-loss sign test \cite{demvsar2006statistical} based on analysis and statistical tests for an overall comparison of the models listed in Table \ref{tab:GE-IFRVFL-CIL}.\\
\textbf{Average AUC:} Using table \ref{tab:GE-IFRVFL-CIL}, the average AUC of the existing models, represented as (AUC, model name), are ($80.76\%$, IFTWSVM), ($70.41\%$, IFKRR), ($79.71\%$, ELM), ($80.31\%$, RVFL), ($82.97\%$, IFRVFL) and ($81.79\%$, GE-IFWRVFL). Whereas the AUC values of proposed GE-IFRVFL-CIL-1 and GE-IFRVFL-CIL-2 models are $84.54\%$ and $84.97\%$, better than that of the existing baseline models. The results show that the proposed models, GE-IFRVFL-CIL-1 and GE-IFRVFL-CIL-2, have the first and second spots in terms of average AUC and the RVFL family models', {\em{i.e.,}} IFRVFL, GE-IFWRVFL, and RVFL models took third, fourth and fifth spots, respectively, followed by ELM, IFTWSVM, and IFKRR. 
\begin{landscape}
\begin{table}[ht]
 \centering
 \caption{\vspace{0.1mm}The classification accuracies of the proposed GE-IFRVFL-CIL-1 and GE-IFRVFL-CIL-2 models along with the existing models, {\em{i.e.,}} IFTWSVM, IFKRR, ELM, RVFL, IFRVFL and GE-IFWRVFL on KEEL \cite{derrac2015keel} imbalanced benchmark datasets.}
 \resizebox{23cm}{!}{
\begin{tabular}{lcccccccc}
\hline
Dataset&IFTWSVM \cite{rezvani2019intuitionistic} &IFKRR \cite{hazarika2021intuitionistic}&ELM \cite{huang2006extreme}&RVFL \cite{pao1994learning}&IFRVFL \cite{malik2022alzheimer}&GE-IFWRVFL \cite{MalikGraph2022}& GE-IFRVFL-CIL-1$^{\dagger}$&GE-IFRVFL-CIL-2$^{\dagger}$ \\
 &(AUC,$C_{1}$)	&(AUC, $C$)		&(AUC, $h_l$)		&(AUC, $h_l$)		&(AUC, $h_l$)&(AUC, $h_l$)&(AUC, $h_l$)&(AUC, $h_l$)\\
    &($C_{3}$, $\mu$)	&($\mu$)	&($C$)	&($C$)	&($C$,$\mu$)	&($C$,$\mu$,$\theta$)&($C$,$\mu$,$\theta$)&($C$,$\mu$,$\theta$)\\
\hline
abalone9-18&$(0.7502, 0.001)$&$(0.7022, 10)$&$(0.6614, 83)$&$(0.7035, 183)$&$(0.7683, 3)$&$(0.8272, 23)$&$(\textbf{0.8319}, 23)$&$(0.8272, 63)$\\&$(10, 0.25)$&$0.125$&$100$&$100$&$(100000, 0.25)$&$(0.1, 0.25, 0.0001)$&$(10000, 0.125, 0.001)$&$(100000, 0.03125, 10)$\\
aus&$(0.8482, 1000)$&$(0.8023, 0.00001)$&$(0.8484, 203)$&$(0.8492, 43)$&$(0.8526, 43)$&$(0.8449, 43)$&$(0.8449, 43)$&$(\textbf{0.8545}, 43)$\\&$(10, 4)$&$0.125$&$0.01$&$0.001$&$(0.001, 32)$&$(0.001, 8, 0.00001)$&$(10, 1, 0.00001)$&$(0.001, 16, 0.00001)$\\
checkerboard\_Data&$(0.8482, 1000)$&$(0.8023, 0.00001)$&$(0.8484, 203)$&$(0.8492, 43)$&$(0.8526, 43)$&$(0.8449, 43)$&$(0.8449, 43)$&$(\textbf{0.8545}, 43)$\\&$(10, 4)$&$0.125$&$0.01$&$0.001$&$(0.001, 32)$&$(0.001, 8, 0.00001)$&$(10, 1, 0.00001)$&$(0.001, 16, 0.00001)$\\
crossplane150&$(\textbf{1}, 0.00001)$&$(0.5873, 10000)$&$(0.9893, 43)$&$(0.9643, 63)$&$(0.9821, 63)$&$(0.9821, 63)$&$(0.9821, 63)$&$(\textbf{1}, 63)$\\&$(0.00001, 0.03125)$&$0.03125$&$100000$&$100$&$(1000, 4)$&$(0.1, 4, 0.1)$&$(0.1, 4, 0.1)$&$(0.1, 8, 1)$\\
ecoli-0-1\_vs\_5&$(\textbf{0.8868}, 100)$&$(0.8259, 0.01)$&$(0.8303, 23)$&$(0.8259, 63)$&$(0.8333, 63)$&$(0.8184, 143)$&$(0.8259, 23)$&$(0.8259, 23)$\\&$(0.01, 0.5)$&$0.25$&$1000$&$0.1$&$(0.1, 8)$&$(1000, 8, 0.00001)$&$(1, 1, 1)$&$(0.00001, 0.25, 0.1)$\\
ecoli-0-1-4-6\_vs\_5&$(0.9321, 100000)$&$(0.9938, 0.00001)$&$(0.9963, 143)$&$(\textbf{1}, 63)$&$(0.9753, 63)$&$(\textbf{1}, 63)$&$(\textbf{1}, 63)$&$(\textbf{1}, 163)$\\&$(0.01, 0.5)$&$0.0625$&$100$&$1000$&$(10, 0.125)$&$(0.00001, 1, 0.001)$&$(0.01, 2, 0.001)$&$(0.001, 2, 0.01)$\\
ecoli-0-3-4-6\_vs\_5&$(0.8083, 10)$&$(0.825, 0.00001)$&$(0.795, 23)$&$(\textbf{0.8333}, 203)$&$(\textbf{0.8333}, 63)$&$(0.825, 43)$&$(\textbf{0.8333}, 43)$&$(0.825, 43)$\\&$(0.001, 1)$&$0.125$&$0.1$&$0.01$&$(10, 2)$&$(0.00001, 2, 0.001)$&$(0.01, 32, 0.1)$&$(0.001, 4, 0.1)$\\
ecoli-0-6-7\_vs\_5&$(0.7336, 0.0001)$&$(0.7172, 0.1)$&$(0.8169, 23)$&$(0.75, 163)$&$(0.7923, 63)$&$(0.8087, 63)$&$(\textbf{0.8675}, 63)$&$(0.8347, 63)$\\&$(0.1, 0.5)$&$0.125$&$10$&$100$&$(0.1, 2)$&$(0.00001, 2, 0.001)$&$(100000, 0.25, 0.001)$&$(10000, 0.25, 0.1)$\\
ecoli0137vs26&$(\textbf{0.9872}, 100000)$&$(0.9321, 1)$&$(0.9529, 43)$&$(0.9706, 143)$&$(0.9449, 123)$&$(0.9706, 203)$&$(0.9578, 63)$&$(0.9706, 183)$\\&$(1000, 4)$&$0.125$&$0.01$&$1$&$(0.01, 1)$&$(10, 1, 0.0001)$&$(10000, 0.03125, 0.00001)$&$(0.0001, 2, 0.01)$\\
glass2&$(0.8065, 0.00001)$&$(0.5, 100000)$&$(0.7121, 83)$&$(0.8105, 83)$&$(0.8145, 3)$&$(0.5565, 203)$&$(\textbf{0.8387}, 143)$&$(0.7258, 43)$\\&$(0.001, 32)$&$8$&$100$&$10000$&$(10000, 0.5)$&$(1, 0.03125, 0.0001)$&$(0.001, 4, 0.0001)$&$(10000, 0.5, 10)$\\
glass5&$(0.4922, 0.01)$&$(0.7031, 0.01)$&$(0.7453, 143)$&$(0.7188, 83)$&$(0.7188, 83)$&$(0.7266, 83)$&$(0.7813, 123)$&$(\textbf{0.9766}, 123)$\\&$(0.001, 4)$&$0.25$&$1$&$1000$&$(100, 8)$&$(100000, 0.25, 0.00001)$&$(0.01, 16, 1)$&$(100000, 0.03125, 1)$\\
haber&$(0.5588, 1000)$&$(0.5891, 1000)$&$(0.5822, 143)$&$(0.5253, 43)$&$(0.6097, 63)$&$(\textbf{0.6138}, 3)$&$(0.6085, 3)$&$(0.5938, 3)$\\&$(0.1, 0.125)$&$8$&$100$&$10$&$(1000, 0.5)$&$(10000, 0.0625, 0.01)$&$(100000, 0.0625, 0.001)$&$(100000, 0.125, 0.1)$\\
heart-stat&$(0.8398, 10)$&$(0.7866, 0.01)$&$(0.8653, 183)$&$(0.8574, 123)$&$(0.8647, 123)$&$(0.8611, 3)$&$(0.8505, 3)$&$(\textbf{0.886}, 103)$\\&$(10, 16)$&$0.5$&$0.0001$&$0.0001$&$(100000, 0.125)$&$(0.001, 4, 0.00001)$&$(0.00001, 4, 0.00001)$&$(10000, 0.03125, 0.00001)$\\
led7digit-0-2-4-5-6-7-8-9\_vs\_1&$(0.9016, 10000)$&$(0.5, 0.00001)$&$(0.9073, 63)$&$(\textbf{0.9419}, 203)$&$(0.9258, 143)$&$(0.9258, 3)$&$(0.9339, 23)$&$(0.9258, 103)$\\&$(100, 2)$&$0.03125$&$100$&$0.01$&$(0.1, 4)$&$(0.1, 2, 0.0001)$&$(0.01, 0.03125, 0.00001)$&$(0.001, 2, 0.01)$\\
monk2&$(\textbf{0.8706}, 100)$&$(0.5, 1)$&$(0.7993, 203)$&$(0.7961, 123)$&$(0.5593, 163)$&$(0.7918, 123)$&$(0.8059, 123)$&$(0.7778, 123)$\\&$(0.00001, 8)$&$0.125$&$10$&$10000$&$(100000, 1)$&$(100000, 2, 1)$&$(10000, 1, 10000)$&$(10000, 1, 10000)$\\
new-thyroid1&$(0.9444, 0.0001)$&$(0.9825, 10)$&$(0.9947, 43)$&$(\textbf{1}, 23)$&$(0.9825, 23)$&$(\textbf{1}, 23)$&$(\textbf{1}, 23)$&$(\textbf{1}, 23)$\\&$(0.0001, 4)$&$0.0625$&$10$&$10$&$(10, 0.03125)$&$(100, 0.5, 0.001)$&$(0.01, 4, 0.001)$&$(0.001, 4, 0.1)$\\
pima&$(0.7582, 0.001)$&$(0.6886, 1)$&$(0.73, 83)$&$(0.7452, 23)$&$(0.7827, 3)$&$(\textbf{0.8048}, 23)$&$(0.8035, 3)$&$(0.787, 23)$\\&$(1, 2)$&$0.03125$&$0.001$&$100$&$(1000, 0.5)$&$(1, 1, 0.001)$&$(100, 0.5, 0.00001)$&$(100000, 0.25, 10)$\\
shuttle-6\_vs\_2-3&$(0.75, 0.0001)$&$(0.9394, 0.00001)$&$(0.9985, 23)$&$(\textbf{1}, 3)$&$(0.9924, 3)$&$(\textbf{1}, 3)$&$(\textbf{1}, 3)$&$(\textbf{1}, 3)$\\&$(0.0001, 2)$&$0.03125$&$10$&$0.1$&$(0.1, 2)$&$(0.1, 8, 0.00001)$&$(0.00001, 0.03125, 0.00001)$&$(0.00001, 0.03125, 10)$\\
shuttle-c0-vs-c4&$(\textbf{1}, 0.0001)$&$(\textbf{1}, 0.00001)$&$(0.9892, 23)$&$(0.9865, 43)$&$(0.9865, 23)$&$(0.9865, 3)$&$(0.9865, 3)$&$(\textbf{1}, 3)$\\&$(0.00001, 16)$&$2$&$0.1$&$0.1$&$(0.0001, 0.5)$&$(0.1, 0.25, 0.00001)$&$(0.001, 4, 0.00001)$&$(0.01, 2, 100)$\\
transfusion&$(0.6146, 0.001)$&$(0.5, 0.00001)$&$(0.606, 123)$&$(0.6257, 23)$&$(0.6555, 163)$&$(0.6382, 163)$&$(0.6245, 63)$&$(\textbf{0.6672}, 23)$\\&$(0.1, 8)$&$0.03125$&$0.1$&$100$&$(100000, 0.25)$&$(1, 0.125, 0.0001)$&$(0.1, 8, 0.0001)$&$(0.1, 32, 0.01)$\\
vehicle1&$(0.7859, 0.001)$&$(0.7366, 1)$&$(0.7626, 163)$&$(0.7683, 83)$&$(0.7954, 203)$&$(\textbf{0.8367}, 163)$&$(0.8339, 163)$&$(0.7997, 83)$\\&$(0.1, 16)$&$0.5$&$100$&$10$&$(100000, 0.03125)$&$(0.1, 1, 0.0001)$&$(10, 1, 0.0001)$&$(100000, 0.5, 0.1)$\\
vowel&$(\textbf{0.9482}, 0.00001)$&$(0.8279, 0.00001)$&$(0.7582, 103)$&$(0.8704, 143)$&$(0.8631, 163)$&$(0.8333, 143)$&$(0.8889, 143)$&$(0.8704, 143)$\\&$(0.001, 1)$&$0.0625$&$100$&$100$&$(100, 2)$&$(1, 2, 0.1)$&$(10, 1, 0.001)$&$(1000, 0.5, 10000)$\\
wpbc&$(0.564, 0.01)$&$(0.5195, 0.1)$&$(0.5811, 143)$&$(0.6716, 3)$&$(\textbf{0.7356}, 203)$&$(0.6002, 163)$&$(0.6252, 163)$&$(0.6558, 3)$\\&$(0.1, 32)$&$1$&$0.1$&$10$&$(0.00001, 0.5)$&$(1000, 1, 0.00001)$&$(0.1, 4, 0.0001)$&$(1000, 2, 10)$\\
yeast-0-3-5-9\_vs\_7-8&$(0.5574, 0.01)$&$(0.5926, 0.1)$&$(0.6126, 163)$&$(0.4926, 203)$&$(0.5963, 163)$&$(0.5444, 143)$&$(\textbf{0.7}, 43)$&$(0.5981, 3)$\\&$(1000, 16)$&$0.25$&$100000$&$10000$&$(1000, 0.5)$&$(0.01, 2, 0.001)$&$(0.00001, 1, 0.0001)$&$(0.1, 1, 0.001)$\\
yeast-0-5-6-7-9\_vs\_4&$(0.7137, 100)$&$(0.6619, 10)$&$(0.6517, 123)$&$(0.6625, 163)$&$(0.7277, 203)$&$(0.7108, 43)$&$(0.7067, 3)$&$(\textbf{0.7775}, 43)$\\&$(10, 16)$&$0.125$&$10$&$1$&$(10000, 0.25)$&$(10, 1, 0.01)$&$(100, 0.5, 0.00001)$&$(10, 0.5, 0.01)$\\
yeast-2\_vs\_4&$(\textbf{0.9507}, 0.01)$&$(0.5, 100000)$&$(0.8172, 103)$&$(0.7822, 63)$&$(0.915, 43)$&$(0.8109, 23)$&$(0.8581, 203)$&$(0.9326, 63)$\\&$(0.1, 16)$&$0.125$&$1$&$100$&$(1000, 1)$&$(100000, 0.5, 0.1)$&$(0.001, 8, 0.001)$&$(100000, 0.03125, 100)$\\
yeast3&$(0.8175, 10000)$&$(0.5, 0.00001)$&$(0.8124, 203)$&$(0.8192, 123)$&$(0.9031, 163)$&$(0.912, 183)$&$(0.9034, 163)$&$(\textbf{0.9298}, 203)$\\&$(10000, 32)$&$0.0625$&$10$&$10$&$(10, 0.03125)$&$(0.01, 0.5, 0.00001)$&$(0.0001, 4, 0.00001)$&$(100, 0.03125, 0.001)$\\
yeast5&$(0.9453, 10000)$&$(0.5, 100000)$&$(0.6542, 143)$&$(0.6655, 163)$&$(\textbf{0.9681}, 203)$&$(0.8251, 183)$&$(0.9335, 23)$&$(0.8942, 43)$\\&$(10000, 16)$&$0.03125$&$1$&$100$&$(1, 0.03125)$&$(100, 1, 0.00001)$&$(0.001, 4, 0.00001)$&$(1000, 0.5, 1)$\\ \hline
Average AUC           & 0.8076 &	0.7041	& 0.7971 &	0.8031 &	0.8297 &	0.8179	& \underline{0.8454} & \textbf{0.8497} \\ \hline
Average Rank               & 4.8036 &	7.0714 &	5.2857 &	4.625 &	3.875	& 4.2679 &	\underline{3.3393}	& \textbf{2.7321}   \\\hline
\end{tabular}}\\
\footnotesize{The boldface in each row signifies the performance of the best model. The underline in the last two rows signifies the second-best model in terms of average AUC and rank. $\dagger$ represents the proposed models.}
\label{tab:GE-IFRVFL-CIL}
\end{table}
\end{landscape}
\hspace{-5mm} The proposed GE-IFRVFL-CIL-2 and GE-IFRVFL-CIL-1 models showed around $14-15\%$ greater AUC than that of IFKRR. Thus, the overall results demonstrate the superiority of the proposed algorithms.

\hspace{-5mm} \textbf{Ranking scheme:}  Although average AUC can be a flawed metric, superior performance in one dataset may compensate for inferior performance in others. To keep this flaw in mind, the models are ranked individually for each dataset to evaluate their respective performances. In this ranking scheme \cite{demvsar2006statistical}, each model is assessed based on its performance on individual datasets: the worst-performing model receives a higher rank, and the best-performing model is ranked lower. Suppose $d$ models are being evaluated using $K$ datasets, and the $l^{th}$ model's rank on the $k^{th}$ dataset is denoted by $\mathcal{r}^k_l$. The $l^{th}$ model's average (overall) rank is determined by the following calculation: $\mathcal{R}_l=\frac{\sum_{k=1}^{K}\mathcal{r}^k_l}{K}$.
The average rank of the models, represented as (average rank, model name), are ($3.3393$, GE-IFRVFL-CIL-1), ($4.2679$, GE-IFWRVFL), ($2.7321$, GE-IFRVFL-CIL-2), ($4.8036$, IFTWSVM), ($7.0714$, IFKRR), ($5.2857$, ELM), ($4.625$, RVFL) and ($3.875$, IFRVFL). The proposed GE-IFRVFL-CIL-2 model has a lower average rank (secured first place) than the competing models, and the proposed GE-IFRVFL-CIL-1 secured third place; a lower rank represents the better performance of the model.

\hspace{-5mm} \textbf{Friedman test:} The Friedman test \cite{friedman1940comparison} is used to statistically analyze the models. The models' average rank is equal under the null hypothesis, assuming they perform equally. The Friedman test follows the chi-squared distribution ($\chi^2_F$) with $d-1$ degrees of freedom (d.o.f.), where $d$ is the number of models being compared. $\chi^2_F = \frac{12K}{d (d+1)} \left(\sum_{l=1}^{d} \mathcal{R}_l^2 - \frac{d(d+1)^2}{4}\right)$ and $F_F=\frac{(K-1)\chi_F^2}{K(d-1)-\chi_F^2}$, where the distribution of $F_F$ has $(d-1)$ and $(K-1)(d-1)$ d.o.f.. For $d=8$ and $K=28$, we get $\chi^2_F=57.5946$ and $F_F=11.2355$. According to the statistical $F-$distribution table, $F_F (7, 189) = 2.0583$ at $5\%$ level of significance. As $11.2355 > 2.0583$, we reject the null hypothesis. The models differ significantly as a result. 

\hspace{-5mm} \textbf{Nemenyi post hoc test:} Using the Nemenyi post hoc test \cite{demvsar2006statistical}, we determine whether there is a significant difference between the models. The critical difference $(C.D.)$ is given by $C.D. = q_\alpha\left(\sqrt{\frac{d(d + 1)}{6K}}\right)$,
 where $q_\alpha$ is the critical value for the two-tailed Nemenyi test from the distribution table. After calculation, we get $C.D.=1.98.$ The models are deemed to be considerably different if there is a $C.D.$ or greater gap in their average ranks. The average rank differences of proposed GE-IFRVFL-CIL-2 from IFTWSVM, 	IFKRR and ELM are $2.0715$, $4.3393$ and $2.5536$, respectively. Whereas the rank of GE-IFRVFL-CIL-1 differs from IFKRR  is $3.7321$. These differences are more than the $C.D.$; therefore, according to the Nemenyi post hoc test, the proposed models GE-IFRVFL-CIL-1 significantly differ from the IFKRR and GE-IFRVFL-CIL-2  is statistically superior to IFTWSVM, IFKRR and ELM. Nevertheless, the GE-IFRVFL-CIL-1 and GE-IFRVFL-CIL-2 models do not significantly differ from other existing models, as determined by the Nemenyi test. However, it is evident that the proposed GE-IFRVFL-CIL-1 and GE-IFRVFL-CIL-2 outperform all the existing baseline models, as indicated by their superior average rank. The win-tie-loss sign test is discussed in Section IV of the Supplementary materials.
 
 The findings exhibit that the utilization of the weighting technique leads to a marked improvement in the generalization performance of GE-IFRVFL-CIL models on imbalanced datasets. Furthermore, the comprehensive analysis affirms that the generalization performance of the proposed GE-IFRVFL-CIL models is significantly enhanced by the integration of GE frameworks, weighting technique, and IF theory compared to the baseline models.
\begin{table*}[ht]
\centering
\caption{Performance of classification models on different levels of noise.}
\label{tab:AVG_NOISE}
\resizebox{\textwidth}{!}{%
\begin{tabular}{|l|c|cccccccc|}
\hline
Datasets &
  Noise &
  IFTWSVM \cite{rezvani2019intuitionistic} &
  IFKRR \cite{hazarika2021intuitionistic} &
  ELM \cite{huang2006extreme}&
  RVFL \cite{pao1994learning} &
  IFRVFL \cite{malik2022alzheimer}&
  GE-IFWRVFL \cite{MalikGraph2022}&
  GE-IFRVFL-CIL-1$^{\dagger}$ &
  GE-IFRVFL-CIL-2$^{\dagger}$ \\ \hline
\textbf{} & $0\%$  & 0.8896 & 0.7747 & 0.8002 & 0.7862 & 0.8862 & 0.8534 & \underline{0.8949} & \textbf{0.9064} \\
\textbf{}        & $5\%$  & 0.8835 & 0.8022 & 0.7985 & 0.7866 & 0.8833 & 0.8576 & \underline{0.8986} & \textbf{0.9038} \\
Average AUC       & $10\%$ & 0.8864 & 0.8009 & 0.7989 & 0.7913 & 0.8895 & 0.8691 & \underline{0.8948} & \textbf{0.9073} \\
\textbf{}        & $20\%$ & 0.8907 & 0.7967 & 0.7994 & 0.8004 & 0.89   & 0.8785 & \underline{0.8939} & \textbf{0.9116} \\
\textbf{}        & $30\%$ & 0.8926 & 0.7917 & 0.7961 & 0.7983 & 0.8907 & 0.8813 & \underline{0.8997} & \textbf{0.9134} \\ \hline
Overall Average AUC&
  \multicolumn{1}{l}{} &
  0.785 &
  0.708 &
  0.7442 &
  0.7496 &
  0.7955 &
  0.7899 &
  \underline{0.8098} &
  \textbf{0.8176}
  \\ \hline
Overall Average Rank &
  \multicolumn{1}{l}{} &
4.04&	6.78&	5.97&	5.4&	4.1	& 4.06&	\underline{3.01}&	\textbf{2.64}
 \\ \hline
\end{tabular}%
}
\footnotesize{The boldface and underline in each row signify the performance of the best and second-best models, respectively. $\dagger$ represent the proposed models.}
\end{table*}
 \begin{table*}
    \centering
    \caption{\vspace{0.1mm}Performance of proposed as well as existing models on ADNI Data.}
    \resizebox{18cm}{!}{
    \begin{tabular}{lcccccccc}
    \hline
    Subjects&IFTWSVM \cite{rezvani2019intuitionistic} &IFKRR \cite{hazarika2021intuitionistic}&ELM \cite{huang2006extreme}&IFRVFL \cite{malik2022alzheimer}&RVFL \cite{pao1994learning}&GE-IFWRVFL \cite{MalikGraph2022}&GE-IFRVFL-CIL-1$^{\dagger}$&GE-IFRVFL-CIL-2$^{\dagger}$\\
    &(AUC, Seny.)	&(AUC, Seny.)		&(AUC, Seny.)		&(AUC, Seny.)		&(AUC, Seny.)&(AUC, Seny.)&(AUC, Seny.)&(AUC, Seny.)	\\
    &(Spey., Pren.)	&(Spey., Pren.)	&(Spey., Pren.)	&(Spey., Pren.)	&(Spey., Pren.)	&(Spey., Pren.)&(Spey., Pren.)&(Spey., Pren.)\\
    \hline
CN\_vs\_AD&$(0.8801,0.9245)$&$(0.7737,0.6981)$&$(0.8444,0.834)$&$(0.898,0.9057)$&$(0.8903,0.8491)$&$(\textbf{0.9049},0.9057)$&$(0.8912,0.9057)$&$(0.8655,0.8679)$\\
&$(0.8356,0.8033)$&$(0.8493,0.7708)$&$(0.8548,0.8083)$&$(0.8904,0.8571)$&$(0.9315,0.9)$&$(0.9041,0.8727)$&$(0.8767,0.8421)$&$(0.863,0.8214)$\\
CN\_vs\_MCI&$(0.6623,0.5902)$&$(0.6329,0.8361)$&$(0.6318,0.5246)$&$(0.6763,0.6885)$&$(0.6295,0.5246)$&$(0.681,0.7213)$&$(0.6986,0.6393)$&$(\textbf{0.7122},0.7213)$\\
&$(0.7344,0.5143)$&$(0.4297,0.4113)$&$(0.7391,0.4894)$&$(0.6641,0.4941)$&$(0.7344,0.4848)$&$(0.6406,0.4889)$&$(0.7578,0.5571)$&$(0.7031,0.5366)$\\
MCI\_vs\_AD&$(0.6212,0.4923)$&$(0.5787,0.3538)$&$(0.5858,0.3662)$&$(\textbf{0.6813},0.8)$&$(0.6621,0.5385)$&$(0.6805,0.7538)$&$(\textbf{0.6813},0.8)$&$(0.6681,0.6308)$\\
&$(0.75,0.5333)$&$(0.8036,0.5111)$&$(0.8054,0.5263)$&$(0.5625,0.5149)$&$(0.7857,0.5932)$&$(0.6071,0.5269)$&$(0.5625,0.5149)$&$(0.7054,0.5541)$\\
\hline
Average AUC&$0.7212$&$	0.6618$&$	0.6873$&$	0.7519$&$	0.7273$&$\underline{0.7555}$&$	\bm{0.757}$&$	0.7486$\\
\hline
\end{tabular}
}
\\\footnotesize{Here, AUC, Seny., Spey. and Pren. denote the area under the curve, sensitivity, specificity and precision, respectively. The boldface and underline in each row signify the performance of the best and second-best models, respectively, in terms of AUC value. $\dagger$ represent the proposed models.}
\label{tab:AD_All_datasets}
\end{table*}
\subsection{Evaluation on KEEL Datasets with Gaussian Noise}
To assess the robustness of the proposed GE-IFRVL-CIL-1 and GE-IFRVFL-CIL-2 models against noise, we contaminate samples with Gaussian noise at levels of $5\%$, $10\%$, $20\%$, and $30\%$ to perturb the features of datasets. These datasets exhibit diverse characteristics in terms of nature and imbalance ratios (IR). A detailed discussion of the selection of these datasets is expounded upon in Section V of the Supplementary material. The detailed experimental results are shown in Supplementary Table IV, and the average AUC of each model \textit{w.r.t.} different level of noise is presented in Table \ref{tab:AVG_NOISE}. 

Upon scrutinizing Table \ref{tab:AVG_NOISE}, it is evident that the proposed GE-IFRVFL-CIL-1 and GE-IFRVFL-CIL-2 models consistently outperformed the baseline models across various noise levels, as indicated by their superior average AUC. Notably, the GE-IRVFL-CIL-2 and GE-IFRVFL-CIL-1 emerged as the top and second-top performers, respectively, at each noise level. With overall average accuracies of $81.76$ and $80.98$, the GE-IFRVFL-CIL-2 and GE-IFRVFL-CIL-1 models secured positions among the top two performers. The IFKRR model demonstrated an overall average AUC of $70.8$, revealing a substantial approximately $10\%$ performance gap when compared to the proposed GE-IFRVFL-CIL models. This stark contrast demonstrates the superior noise-handling capabilities of the proposed models. The statistical measure values F-Measure and G-Means are discussed in the Supplementary Section III, and their corresponding graph is also presented in Fig. S1(a) and S1(b) \textit{w.r.t.} varying percentage ($~5\%, ~10\%, ~20\% ~\text{and} ~30\%$) of the existing as well as the proposed GE-IFRVFL-CIL-1 and GE-IFRVFL-CIL-2 models.

Further, the average ranks affirm the excellence of the GE-IFRVFL-2 and GE-IFRVFL-1 models, standing at $2.64$ and $3.01$, respectively. These ranks represent the lowest and second lowest among all compared models. This statistical evidence substantiates the robustness of the proposed GE-IFRVFL-CIL models in the presence of noise, consistently surpassing baseline models in a statistically significant manner.
\subsection{Evaluation on ADNI Dataset}
Alzheimer's disease (AD) is a brain dysfunction illness that gradually impairs people's memory and cognitive abilities. In order to train the proposed models GE-IFRVFL-CIL-1 and GE-IFRVFL-CIL-2, we use scans from the Alzheimer's Disease Neuroimaging Initiative (ADNI) dataset, which is accessible at $adni.loni.usc.edu$. The ADNI project was started in 2003 with the intention of examining neuroimaging techniques such as magnetic resonance imaging (MRI), positron emission tomography (PET), and other diagnostic tests of AD at the stage of mild cognitive impairment (MCI). The feature extraction pipeline is the same as the methodology outlined in \cite{richhariya2020diagnosis}. The dataset consists of three cases: control normal (CN) versus moderate cognitive impairment (MCI) (CN\_vs\_MCI), MCI versus AD (MCI\_vs\_AD), and  CN versus AD (CN\_vs\_AD).

The AUC values of the proposed GE-IFRVFL-CIL models as well as existing models for AD diagnosis are shown in Table \ref{tab:AD_All_datasets}. We analyze that with an average AUC of $75.70\%$, the proposed GE-IFRVFL-CIL-1 is the best classifier. The AUC of the remaining models, IFTWSVM, IFKRR, ELM, RVFL, IFRVFL GE-IFWRVFL, and GE-IFRVFL-CIL-2, are $72.12\%$, $66.18\%$, $68.73\%$, $72.73\%$, $75.19\%$, $75.55\%$ and $74.86\%$, respectively. In comparison to the AUC of IFKRR, the proposed models (GE-IFRVFL-CIL-1 and GE-IFRVFL-CIL-2) have an average AUC that is around $9.52\%$ and $8.68\%$ higher, respectively.
With an AUC of $71.22\%$ in the CN\_vs\_MCI case, the proposed GE-IFRVFL-CIL-2 came out on top followed by the proposed GE-IFRVFL-CIL-1, which has an AUC of $69.86\%$. 
 With $68.13\%$ AUC value, the proposed GE-IFRVFL-CIL-1 model and the existing IFRVFL model are the most accurate classifiers for the MCI\_vs\_AD comparison.
 Therefore, the proposed GE-IFRVFL-CIL-1 model has an overall winning performance, whereas the proposed GE-IFRVFL-CIL-2 wins in the CN\_vs\_MCI subject and comes under the top four models in terms of overall performance. Of all models, IFKRR has the lowest average AUC, followed by ELM. The top five models (in terms of average AUC) across all the compared models are from the RVFL family, demonstrating the superior generalization capabilities and dominance of the RVFL-based models. The analysis of F-measure and G-mean is conducted in Section VI of the Supplementary material.
 \section{Conclusion and Future Work}
 \label{conclusion}
The conundrum of class imbalance poses a formidable challenge for the RVFL model, thereby impeding its efficacy in accurately classifying minority classes. 
We propose novel GE-IFRVFL-CIL-1 and GE-IFRVFL-CIL-2 models to address the predicament of imbalance classification problems. Both the proposed models leverage graph embedding, intuitionistic fuzzy theory, and a class weighting mechanism. As a result, the proposed models handle CI issues, noisy samples, and outliers in the datasets while simultaneously conserving the inherent geometric structure of the data. The efficacy of the proposed models for class imbalance learning is demonstrated through their application to $28$ standard benchmark datasets sourced from the KEEL imbalanced dataset repository across a range of diverse domains. Statistical evaluations, including the average AUC, ranking scheme, Friedman test, Nemenyi post hoc test, and win-tie-loss sign test, showed that the proposed GE-IFRVFL-CIL-1 and -2 models outperform other classifiers. Moreover, we contaminate the dataset's features with Gaussian noise; the proposed GE-IFRVFL-CIL-2 and GE-IFRVFL-CIL-1 models exhibit the highest and second-best performance, respectively. In order to demonstrate the efficacy and generic nature of the proposed GE-IFRVFL-CIL-1 and GE-IFRVFL-CIL-2 models, we apply them to the ADNI dataset for the purpose of diagnosing Alzheimer's disease. Upon analysis, it is determined that the proposed GE-IFRVFL-CIL-1 model exhibits the highest degree of classification performance as measured by the average AUC metric.  As per the findings of the research \cite{tanveer2020machine}, it was concluded that the MCI versus AD subject represents the most challenging issue to be addressed in the context of diagnosing Alzheimer's disease. Notwithstanding, the proposed GE-IFRVFL-CIL-1 model achieves the foremost position in this scenario, thus exhibiting its superiority over other models. The proposed model GE-IFRVFL-CIL-2 turned out to be the most effective in diagnosing CN\_vs\_MCI subjects, closely followed by the GE-IFRVFL-CIL-1 model. In this paper, our study focused on shallow RVFL, which has limited potential for feature representation learning. Our forthcoming plan is to expand this research to deep and ensemble variations of RVFL. The proposed models exhibit a slightly higher computational complexity than the RVFL model. Consequently, reducing computational complexity while preserving efficacy will be one of the main research goals. The current study is limited to binary classification; however, extending the proposed model to multi-class classification is an interesting avenue for future research. The code of the proposed GE-IFRVFL-CIL is available at https://github.com/mtanveer1/GE-IFRVFL-CIL.
 \section*{Acknowledgment}
This project is supported by the Indian government's Department of Science and Technology (DST) and Ministry of Electronics and Information Technology (MeitY) through two grants: DST/NSM/R\&D\_HPC\_Appl/2021/03.29 for the National Supercomputing Mission and MTR/2021/000787 for the Mathematical Research Impact-Centric Support (MATRICS) scheme.
The Council of Scientific and Industrial Research (CSIR), New Delhi, provided a fellowship for Md Sajid's research under the grants 09/1022(13847)/2022-EMR-I. The authors appreciate the facilities and assistance offered by IIT Indore. The dataset employed in this study was procured with the aid of financial support from the Alzheimer's Disease Neuroimaging Initiative (ADNI), which was made possible through the National Institutes of Health's U01 AG024904 grant and the Department of Defense's ADNI award W81XWH-12-2-0012. The aforementioned initiative's funding was sourced from the National Institute on Aging, the National Institute of Biomedical Imaging and Bioengineering, and several munificent contributions made by a variety of entities: F. Hoffmann-La Roche Ltd. and its affiliated company Genentech, Inc.; Bristol-Myers Squibb Company;  Alzheimer’s Drug Discovery Foundation; Merck \& Co., Inc.;  CereSpir, Inc.; Meso Scale Diagnostics, LLC.;  Novartis Pharmaceuticals Corporation; AbbVie, Alzheimer’s Association; Lumosity; Biogen; Fujirebio; IXICO Ltd.; Araclon Biotech; BioClinica, Inc.; NeuroRx Research; EuroImmun; Piramal Imaging; GE Healthcare; Cogstate; Eisai Inc.; Johnson \& Johnson Pharmaceutical Research \& Development LLC.;  Servier; Eli Lilly and Company; Transition Therapeutics Elan Pharmaceuticals, Inc.; Janssen Alzheimer Immunotherapy Research \& Development, LLC.; Lundbeck; Neurotrack Technologies; Pfizer Inc. and Takeda Pharmaceutical Company. Financial aid from the Canadian Institutes of Health Research is being extended to sustain ADNI clinical sites across Canada. Meanwhile, the Foundation for the National Institutes of Health (www.fnih.org) has facilitated private sector donations to support this endeavor.  Support for the grants dedicated to research and education was furnished by the Northern California Institute and the Alzheimer's Therapeutic Research Institute at the University of Southern California. The data associated with the ADNI initiative were made available through the auspices of the Neuro Imaging Laboratory located at the University of Southern California. The present study relied upon the ADNI dataset, which can be accessed via adni.loni.usc.edu. The ADNI initiative was planned and executed by the ADNI investigators, although they did not contribute to either the analysis or writing of this particular article. A thoroughly detailed listing of ADNI investigators can be accessed via the following link: \url{http://adni.loni.usc.edu/wp-content/uploads/how_to_apply/ADNI_Acknowledgment_List.pdf}.
\bibliographystyle{IEEEtranN}
 \bibliography{refs}
\end{document}